\documentclass[11pt]{article}

\usepackage[margin=1in]{geometry}
\usepackage[T1]{fontenc}
\usepackage[utf8]{inputenc}
\usepackage{lmodern}
\usepackage{microtype}
\usepackage{amsmath,amssymb}
\usepackage{booktabs}
\usepackage{tabularx}
\usepackage{enumitem}
\usepackage{xcolor}
\usepackage{graphicx}
\usepackage{caption}
\usepackage{tikz}
\usetikzlibrary{arrows.meta,positioning,fit,backgrounds,calc,shapes.geometric}
\usepackage{pgfplots}
\pgfplotsset{compat=1.18}
\usepackage[colorlinks=true,linkcolor=blue!50!black,citecolor=blue!50!black,urlcolor=blue!50!black]{hyperref}
\usepackage{color-edits}
\addauthor{vs}{red}
\addauthor{jb}{blue}
\addauthor{jt}{orange}

\definecolor{tcblue}{HTML}{2F5D8A}
\definecolor{tcorange}{HTML}{D9822B}
\definecolor{tcgreen}{HTML}{3A8F5C}
\definecolor{tcred}{HTML}{B8423A}
\definecolor{tcgray}{HTML}{7A7A7A}
\definecolor{tcpurple}{HTML}{6B4C9A}

\newcommand{\TC}{\textsc{PaperFormalizer}}

\setlist{itemsep=2pt,topsep=3pt}
\title{\TC:  Author-in-the-Loop Autoformalization of Research Papers in Lean}
\author{Jiyuan Tan \qquad Jose Blanchet\thanks{Joint senior authors, listed alphabetically.} \qquad Vasilis Syrgkanis\footnotemark[1]\\[4pt]
\normalsize Stanford University}
\date{}

\begin{document}
\maketitle

\begin{abstract}
Even a carefully proofread paper can leave small gaps in its proofs or its stated assumptions.
We introduce \TC, a Lean-based agentic pipeline that helps authors find and close such gaps in their own papers and turn each paper into a Lean-verified version.
Starting from the \LaTeX{} source, \TC{} transcribes the paper into a dependency graph of statements, audits each statement in natural language for faithfulness, ambiguity and correctness, and then formalizes and proves the statements in Lean against Mathlib, optional domain libraries and the author's personal toolkit.
Only the author knows what a statement is meant to say and how a formalization gap in a result is best fixed.
The pipeline therefore stops and asks the author whenever a statement admits more than one reading, presents every repair, in natural language and \TeX{} symbols, as a proposal that the author can accept, decline or redirect, and leaves to the author every decision that could potentially change what the paper claims.
The output is an annotated and revised version of the paper, a report of the changes, and Lean code in which each declaration is linked to the statement it formalizes.
The author's toolkit grows as more papers are formalized, and authors can share their verified papers so that researchers who build on them can reuse their Lean code.
By making verified papers practical to produce, \TC{} aims to make research that rests on mathematical proofs more robust.
\TC{} runs inside a coding agent (Claude Code or Codex) and is available at \url{https://github.com/Jiyuan-Tan/PaperFormalizer}. Papers with Lean code shared by users are available at \url{https://paperformalizer.org}. 
\end{abstract}


\section{Introduction}

Many theoretical papers in statistics, machine learning, operations research and mathematics rest on long technical proofs with many tedious details.
Even after several rounds of proofreading, an author can easily overlook a detail that seems unimportant, such as a boundary case, a missing regularity condition, or a step that needs an assumption the paper never states.
Most of the time, such gaps have no effect on the main conclusions.

A proof assistant such as Lean~\cite{lean4}, with its mathematical library Mathlib~\cite{mathlib}, accepts a proof only when every step follows from the definitions and earlier results, so no gap can pass unnoticed.
Formalizing a paper in Lean is therefore a reliable way for authors to find and close the gaps in their own proofs, and the result is a paper whose mathematics is Lean-verified.
Formalizing a research paper by hand, however, can take weeks of work from someone fluent in both the paper's field and Lean.
Recent agentic systems formalize informal mathematics automatically, but a research paper raises two problems that a machine cannot settle alone.
First, natural language is ambiguous. A statement can admit several readings, and a formalization of the wrong one proves something the author never meant.
Second, when a statement turns out to be false, there are usually many ways to fix it. The repair that changes the least in logic may still miss what the author intends, for example by giving up the part of the result the author cares about most.
Both problems depend on the author's intent, so the author has to take part in the process.
We introduce \TC, a Lean-based pipeline that helps authors improve their own papers.
\TC{} points the author to each gap in the proofs, recommends how to close it, and produces a Lean-verified version of the paper, while the author keeps every decision about what the paper claims.

The pipeline has two stages, natural language review and autoformalization (Figure~\ref{fig:pipeline}).
In the first stage, one agent transcribes the \TeX{} source into a dependency graph whose nodes are the statements of the paper.
A second agent then checks each node closely for correctness and ambiguity and proposes repairs for the problems it finds.
Whenever the agent is unsure what a statement is meant to say, it stops and asks the author, {in natural language and \TeX{} symbols}, and the author decides on every repair that would weaken a main result.
Once the paper passes this stage, the pipeline formalizes each statement in Lean automatically.
The final output is a revised paper and a report of what was changed, together with the Lean code.
We used \TC{} on three of our own papers. Five of the seven results in one paper and 15 of the 29 results in another needed some revision, and for each of them the pipeline proposed a repair that {largely preserve the statement of the main results in the paper.} (Section~\ref{sec:case} and Table~\ref{tab:shared}). 

\begin{figure}[t]
\centering
\begin{tikzpicture}[
  font=\footnotesize,
  >={Stealth[length=2.2mm]},
  stage/.style={draw=#1!80!black, fill=#1!10, rounded corners=3pt, line width=0.6pt,
                minimum width=2.7cm, minimum height=1.15cm, text width=2.6cm, align=center},
  io/.style={draw=tcgray, fill=gray!8, rounded corners=3pt, line width=0.6pt,
             minimum width=2.4cm, minimum height=1.15cm, text width=2.3cm, align=center},
  person/.style={draw=tcpurple, fill=tcpurple!10, rounded corners=8pt, line width=0.6pt,
                 minimum width=2.2cm, minimum height=0.9cm, text width=2.1cm, align=center},
  flow/.style={->, line width=0.7pt, color=black!70},
  talk/.style={<->, dashed, line width=0.6pt, color=tcpurple},
  group/.style={draw=#1!50, fill=#1!4, rounded corners=6pt, inner sep=7pt, line width=0.5pt},
]
\node[io]                         (src) at (0,0)    {\textbf{\LaTeX{} source}\\ \scriptsize file, folder or arXiv e-print};
\node[stage=tcblue]               (s1)  at (3.3,0)  {\textbf{1.~Transcription}\\ \scriptsize dependency graph of statements};
\node[stage=tcblue]               (s2)  at (6.6,0)  {\textbf{2. Review}\\ \scriptsize faithfulness, ambiguity, correctness; repairs};
\node[person]                     (usr) at (11.8,0) {\textbf{Author}\\ \scriptsize via a local web page};

\node[stage=tcorange]             (s3)  at (0,-3.3)    {\textbf{3. Planning}\\ \scriptsize reuse, formalize, or keep cited};
\node[stage=tcorange]             (s4)  at (3.3,-3.3)  {\textbf{4. Scaffolding}\\ \scriptsize Lean statements, faithfulness audit};
\node[stage=tcorange]             (s5)  at (6.6,-3.3)  {\textbf{5. Proof filling}\\ \scriptsize with fine-grained faithfulness audit};
\node[stage=tcorange]             (s6)  at (9.9,-3.3) {\textbf{6. Linking}\\ \scriptsize Lean declaration $\leftrightarrow$ paper statement};
\node[io, text width=2.5cm, minimum width=2.6cm] (out) at (13.2,-3.3) {\textbf{Report}\\ \scriptsize annotated \& revised paper, Lean code};

\node[draw=tcgreen!80!black, fill=tcgreen!8, rounded corners=3pt, line width=0.6pt,
      minimum height=0.8cm, text width=6.2cm, align=center] (lib) at (1.7,-5.2)
      {\textbf{Libraries:} Mathlib $\cdot$ domain libraries $\cdot$ personal toolkit};

\begin{scope}[on background layer]
  \node[group=tcblue,   fit=(s1)(s2), label={[font=\footnotesize\bfseries,text=tcblue]above:Stage I: natural language review}] (g1) {};
  \node[group=tcorange, fit=(s3)(s6), label={[font=\footnotesize\bfseries,text=tcorange!80!black,anchor=north east]south east:Stage II: autoformalization}] (g2) {};
\end{scope}

\draw[flow] (src) -- (s1);
\draw[flow] (s1) -- (s2);
\draw[flow] (s2.south) -- ++(0,-0.62) -| node[pos=0.25,above,font=\scriptsize]{accepted, revised statements} (s3.north);
\draw[flow] (s3) -- (s4);
\draw[flow] (s4) -- (s5);
\draw[flow] (s5) -- (s6);
\draw[flow] (s6) -- (out);

\draw[talk] (usr.west) -- node[below,font=\scriptsize,align=center]{readings, repairs,\\ issue reports} (g1.east);
\draw[talk] ($(usr.south)+(-0.35,0)$) -- node[right,font=\scriptsize,align=left,pos=0.45]{plan review,\\ statement\\ revisions} ($(usr.south |- g2.north)+(-0.35,0)$);

\draw[flow, color=tcgreen!70!black] (lib.north -| s3.south) -- node[left,font=\scriptsize]{reuse} (s3.south);
\draw[flow, dashed, color=tcgreen!70!black] (out.south) |- node[pos=0.75,below,font=\scriptsize]{promotion of proved results (author decides)} (lib.east);
\end{tikzpicture}
\caption{Workflow of \TC. Stage I turns the \LaTeX{} source into a reviewed and repaired dependency graph. Stage II formalizes and proves the revised statements in Lean. Dashed purple arrows mark the points where the author takes part (Section~\ref{sec:author}); the dashed green arrow is the feedback loop into the personal toolkit (Section~\ref{sec:toolkit}).}
\label{fig:pipeline}
\end{figure}

Each verified paper also makes the next one cheaper to verify.
A researcher often works in related areas and uses the same techniques across papers, so much of the Lean code written for one paper can be reused in the next.
Each author can keep a personal toolkit that grows as they formalize more papers.
We also provide a place where authors can share a verified paper together with its Lean code, so that researchers who build on the paper can reuse most of that code.

\paragraph{Intended use.}
We designed \TC{} for authors who want to strengthen their own papers, before submission or after publication.
The pipeline makes its key decisions together with the author, because the right reading of a statement and the right repair of an error depend on what the author intends.
A repair proposed without the author is only the pipeline's best guess, and the author may have a different reading in mind or prefer a different fix.
\TC{} is therefore a tool for improving a paper together with its authors.

\paragraph{Related work.}
There is some recent work in this direction.
OSVerify~\cite{osverify} and Rethlas-OSV review the mathematics of solutions and papers in natural language, passing only audited findings to human experts.
Rethlas and Archon~\cite{rethlas}, LeanMarathon~\cite{leanmarathon} and LeanFlow~\cite{leanflow} formalize informal arguments or whole papers in Lean~4.
\TC{} sits between these two scenarios and puts the author in the loop, with the goal of producing a verified version of the author's paper. It first reviews the paper in natural language and proposes repairs for the errors it finds, then formalizes every statement in Lean, and it asks the author whenever a statement can be read in more than one way or a repair would {qualitatively change} a main result.

\section{Pipeline Design}
\label{sec:pipeline}

\begin{figure}[!htb]
\centering
\includegraphics[width=0.72\textwidth]{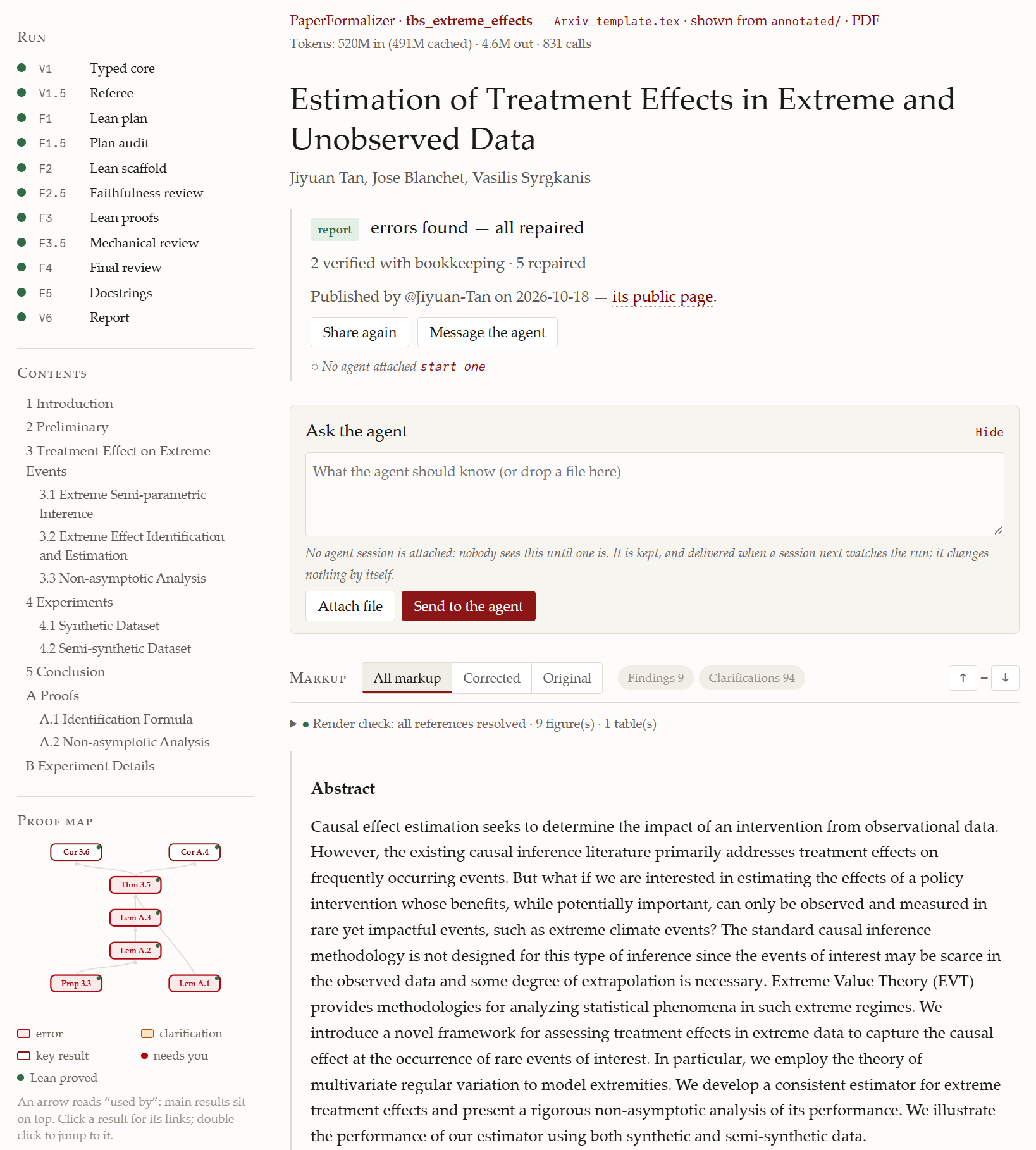}
\caption{The page of the case-study paper (Section~\ref{sec:case}) after its run; the viewer carries the name of the public website, PaperFormalizer. The left column tracks the run's stages (V1 is Stage~1, V1.5 Stage~2, F1--F1.5 Stage~3, F2 Stage~4, F2.5--F4 Stage~5, and F5--V6 Stage~6), gives the paper's contents, and draws the proof map, in which an arrow reads ``used by'' and each result is colored by its review outcome. Above the paper are the report's verdict and the box in which the author writes to the agent. The paper itself follows, with the pipeline's marks; Figures~\ref{fig:ui-finding} and~\ref{fig:ui-lean} show one of its results.}
\label{fig:ui-page}
\end{figure}

The pipeline has six stages.
The first two form the natural language review: an agent transcribes the \TeX{} source into a dependency graph, and a reviewer agent checks the transcription and the ambiguity and correctness of each statement, asking the author for the intended reading when needed.
The remaining four form the formalization: planning, scaffolding, proof filling, and linking.
The author starts and controls the run from a coding agent. The agent also opens a local web page that shows the paper with the pipeline's marks added as the run proceeds, and the author makes each decision there (Figure~\ref{fig:ui-page}). We describe each stage below and the role of the author in Section~\ref{sec:author}.

\paragraph{Stage 1: Transcription.}
The input is a \texttt{.tex} file, a folder, or an archive such as an arXiv e-print.
An agent splits the paper into nodes, one per definition, assumption, lemma, proposition, theorem, corollary, or cited result, and adds an edge whenever a statement or proof uses another node (Figure~\ref{fig:graph}a). If a node's text breaks off inside a formula or an environment, the node is sent back to the transcription agent.

\paragraph{Stage 2: Natural language review.}
A reviewer agent then goes through the graph in dependency order and examines every node along three axes:
faithfulness (does the node match the paper?),
ambiguity (is there exactly one sensible reading?), and
correctness (does the statement hold, and does its proof establish it?).
When the intended reading can be inferred from context, the reviewer pins it. The most plausible reading becomes an explicit assumption that all later stages follow. It appears as a clarification on the paper's page, where the author can edit or remove it.
When the text does not settle the intended meaning, the agent stops and asks the author.
The reviewer reports a result as wrong only if it finds a counterexample within the paper's setting. If every counterexample is a degenerate case that readers would exclude anyway (for example, zero variance or an empty class), the reviewer adds a side condition instead.
For each error, a fixer agent proposes a small repair that keeps the key contribution of the result, and a separate reviewer checks the repair.
A small repair is still a guess at what the author intends, so the author keeps control over it.
A repair that would {substantively change} a main result goes to the author, who can accept it, decline it, ask the fixer to try again with guidance, or withdraw the finding when it misreads the paper.
Every other repair is applied and shown in the annotated paper, with the original in red and the repair in blue, so the author can see each change and report the ones they disagree with.
The reviewer also checks each cited result against its source. It downloads the source itself, or asks the author for it if it cannot, and reports any place where the paper uses a citation for more than the source states.

\begin{figure}[t]
\centering
\begin{minipage}[c]{0.38\textwidth}
\centering
\begin{tikzpicture}[
  font=\scriptsize, >={Stealth[length=1.8mm]},
  nd/.style={draw=#1!80!black, fill=#1!15, rounded corners=2pt, line width=0.6pt,
             minimum width=1.45cm, minimum height=0.6cm, align=center},
  e/.style={->, line width=0.55pt, color=black!60},
]
\node[nd=tcgray]   (c)  at (0,0)     {Cited [7]};
\node[nd=tcgreen]  (d)  at (1.75,0)  {Def.~2.1};
\node[nd=tcorange] (a)  at (3.5,0)   {Assum.~1};
\node[nd=tcgreen]  (l1) at (0.6,-1.2) {Lem.~3.1};
\node[nd=tcred]    (l2) at (2.9,-1.2) {Lem.~3.2};
\node[nd=tcgreen, line width=1.2pt] (t) at (1.75,-2.4) {\textbf{Thm.~3.3}};
\draw[e] (c) -- (l1); \draw[e] (d) -- (l1); \draw[e] (d) -- (l2); \draw[e] (a) -- (l2);
\draw[e] (l1) -- (t); \draw[e] (l2) -- (t);
\node[anchor=north west, align=left, font=\scriptsize] at (-0.75,-3.0) {%
  \tikz\fill[tcgreen!40] (0,0) rectangle (0.22,0.15); ok\quad
  \tikz\fill[tcorange!40] (0,0) rectangle (0.22,0.15); pinned reading\\
  \tikz\fill[tcred!40] (0,0) rectangle (0.22,0.15); error, repaired\quad
  \tikz\fill[gray!40] (0,0) rectangle (0.22,0.15); kept cited};
\node[font=\small\bfseries] at (-0.6,0.65) {(a)};
\end{tikzpicture}
\end{minipage}%
\hfill
\begin{minipage}[c]{0.6\textwidth}
\centering
\small
\textbf{(b)}\par\smallskip
\begin{tabular}{@{}ll@{}}
\toprule
Category & Meaning \\
\midrule
proved & proved in Lean as stated \\
side condition & proved with an added regularity condition \\
conditional & proved from a cited result or an assumption \\
repaired & false as printed, fixed by a minimal repair \\
refuted & shown false, with no confirmed repair \\
not determined & the statement leaves its meaning open \\
unresolved & neither proved nor refuted \\
\bottomrule
\end{tabular}
\end{minipage}
\caption{(a) A toy dependency graph after Stage~2. Each node is a statement of the paper; edges point from a result to the results that use it. Colors show the review outcome. (b) The categories in which the final report places each result of the paper.}
\label{fig:graph}
\end{figure}

\paragraph{Stage 3: Planning.}
Once the reviewed and possibly repaired graph is accepted, a planning agent decides, for each statement, whether to reuse it from an existing library, formalize it, or leave it unformalized.
Reuse draws on Mathlib, on domain libraries chosen at installation (for example, Causalean~\cite{causalsmith} for causal inference, StatLean~\cite{statlean} for statistics, FoML~\cite{foml} for learning theory and optlib~\cite{optlib} for optimization), and on the author's personal toolkit. The agent searches these libraries by keyword and with the Loogle~\cite{loogle} and LeanSearch~\cite{leansearch} search engines.
Cited results that are not among the paper's main results usually stay cited, since formalizing them can be too costly and slow; they enter the Lean code as explicit assumptions and are listed in the report.
A separate agent checks that the plan is faithful to the paper, reuses existing results where it can, and keeps the right results cited. 

\paragraph{Stages 4 and 5: Scaffolding and proof filling.}
Both stages check faithfulness to the paper in fine detail.
Scaffolding writes every planned statement in Lean with its proof left open, compiles the result, and checks each Lean statement against the corresponding natural language statement.
Proof filling then runs as a loop: workers fill the proofs, querying Lean through its language server, and reviewer agents check that each proved statement is faithful to the paper and that its proof uses no placeholder (\texttt{sorry}) and no axioms beyond Lean's standard ones.
If the Lean work shows that a paper statement is wrong as written, the pipeline proposes to return to the review stage and repair that statement, and the author approves or rejects the proposal.

\paragraph{Stage 6: Linking and report.}
Finally, the pipeline links each Lean declaration to the statement of the paper it formalizes, so that the author can easily confirm that each Lean statement says what the paper means.
The final report places each result in one of the categories of Figure~\ref{fig:graph}b, lists every added condition, pinned reading and repair, and tells the author how much of the mathematics needed changes, at one of three levels: major errors, minor issues, or no errors found.
Alongside the report, the pipeline writes an annotated copy of the paper (each finding marked, with the original in red and the revision in blue) and a revised version of the paper, together with the Lean code.

\paragraph{Models.}
Most roles run on OpenAI's \texttt{gpt-6.1-sol} through the Codex CLI, at high or medium reasoning effort; the Lean plan, the second independent reviewer in the proof-review loop, and the curation of the personal toolkit run on Claude Opus through the Claude Code CLI.
With two model families, the plan and the final review come from a model other than the one that writes the Lean code.
When Claude Code is not available, its roles fall back to \texttt{gpt-6.1-sol} at high reasoning effort, and every model can be replaced through an environment variable.

\section{The Author in the Loop}
\label{sec:author}

The author takes part in a run in three ways. They make the decisions that change what the paper claims, they correct the pipeline when it misreads the paper, and they own every output of the run.

\paragraph{Decisions reserved for the author.}
The agent never makes the following decisions on its own: a repair that would {substantively change} a main result, any other ruling that would leave a main result {substantively different} than printed (unless it only adds a bookkeeping assumption such as measurability), the promotion of results into the toolkit, and the sharing of a paper.
Each of these decisions appears on the paper's page with the agent's recommendation and its reason beside it, and the run waits until the author answers.
A ruling the agent makes on its own that leaves a lemma or a definition weaker than printed is listed under ``Review the agent's decisions''. There the author approves it, or rejects it and says what to do instead, and the agent acts on a rejection at once.

\paragraph{Correcting the pipeline.}
The pipeline can misread a paper, and the author is the person best placed to notice.
Every comment on the paper's page, whether a finding, an applied repair or a clarification, has a link to report an issue. The author describes what is wrong, the agent takes it up, and the comment is marked once the issue is addressed.
The author can also edit or remove a pinned reading, add a side condition, and withdraw a finding that misreads the paper.
When the agent needs something only the author has, such as a cited paper it could not download, it asks on the page, and the author can reply with files.
At any time, the author can message the agent, for example to explain what a passage is meant to say.

\paragraph{Ownership of the output.}
The revised paper keeps the author's own class, preamble, bibliography and file layout, with each revised statement and proof written in place.
It is plain \TeX{} with no macro or note of \TC, so the author can keep editing it.
The annotated paper shows every change the run made, proofs included, so the author can compare each change with what they meant.
The Lean code is linked statement by statement to the paper, and the author decides which results enter the toolkit (Section~\ref{sec:toolkit}) and whether the paper is shared.

\section{Personal Toolkit and Paper Sharing}
\label{sec:toolkit}

Each run takes place inside the author's toolkit, a Lean project owned by the author. The project has three tiers. The bottom tier holds the read-only base libraries (Mathlib and the chosen domain libraries). The middle tier holds the author's reusable results, split into Foundations (standard results missing from the base libraries) and Tricks (the author's own reusable lemmas). The top tier holds one module tree per formalized paper.

\paragraph{The feedback loop.}
A run offers results to the toolkit twice.
After planning, it proposes the results the paper cites without proving; an accepted one is built and proved in the toolkit, and the paper then uses it in place of an assumption.
After the report, it proposes the results the paper's Lean has proved; an accepted one moves into the toolkit, or its general form is built there and the paper's result becomes a corollary of it.
Each proposal is written so that the author does not need to read Lean. It gives the statement in ordinary mathematical language, a reference, the reason it is reusable, the places where the author's other papers state it, and the closest existing results.
Nothing enters the toolkit without the author's decision, and Lean checks that a promotion leaves every statement of the paper and the toolkit exactly as it was.
Over time, the toolkit collects the definitions and techniques of the author's own line of work, and each new paper in that line has fewer results to formalize from scratch.

\paragraph{Paper sharing.}
Authors can share a verified paper on our public website, where others can build on it.
A shared paper includes the revised \TeX{} source and PDF, the findings with the reviewer agent's notes, the report, and the Lean code.
Sharing is optional and is always the author's decision. A shared paper is public and appears under the author's name; the author signs in with GitHub only to confirm their identity, and the sign-in requests no access permissions.
A researcher who builds on a shared paper can start from its Lean code and proved results instead of formalizing them again.
Each shared paper adds verified results to a common collection, so later work in the same area has more checked mathematics to build on.

\section{Case Study: Treatment Effects on Extreme Events}
\label{sec:case}

We used \TC{} to produce a Lean-verified version of one of our own papers, \emph{Estimation of Treatment Effects in Extreme and Unobserved Data}. The paper defines the normalized extreme treatment effect $\theta^{\mathrm{NETE}}$, the effect of a treatment $D$ on an outcome $Y$ when an unobserved noise vector $U$ takes extreme values, and models $U$ as multivariate regularly varying with tail index $\beta$.
It identifies $\theta^{\mathrm{NETE}}$ as the product of an angular factor and a radial factor (Proposition~3.3), proposes doubly robust and inverse propensity weighted estimators that use an adaptive Hill estimator for the tail index, and proves finite-sample rates for them (Theorem~3.5 and its corollaries and lemmas).
The run used Mathlib, Causalean, StatLean and the authors' personal toolkit.

\paragraph{What the run produced.}
The run produced a revised version of the paper in which all seven results are proved in Lean, with no \texttt{sorry} and no axioms beyond Lean's standard ones.
To get there, the run found that five of the seven results did not hold as printed and helped us repair each one.
The review traced these errors to two independent roots, Proposition~3.3 and Lemma~A.1 (the rate of the adaptive Hill estimator). Theorem~3.5 and Lemmas~A.2 and~A.3 rest on Proposition~3.3 and inherit its error, and their proofs also contain errors of their own.
For each error the fixer proposed a repair and a second reviewer confirmed it. We read each proposal on the paper's page and, as authors, accepted the weaker statement. The two corollaries had no error of their own and were restated to follow the revised results.
Besides the repairs, the run made 94 readings explicit (Section~\ref{sec:pipeline}, Stage~2) and listed them as clarifications. For example, it fixed that the homogeneity inequality of Assumption~3.2 holds for both treatments and every radius, uniformly over the bounded covariates and the positive unit sphere of the $\ell_1$ norm.
Four further findings, raised while the Lean statements were written, concerned points the paper leaves open, such as whether the overlap constant may depend on the sample size. Only the authors could settle these points. We answered each by stating our reading, the Lean code adopts that reading {and marks the issue as resolved.}

\begin{figure}[t]
\centering
\includegraphics[width=\textwidth]{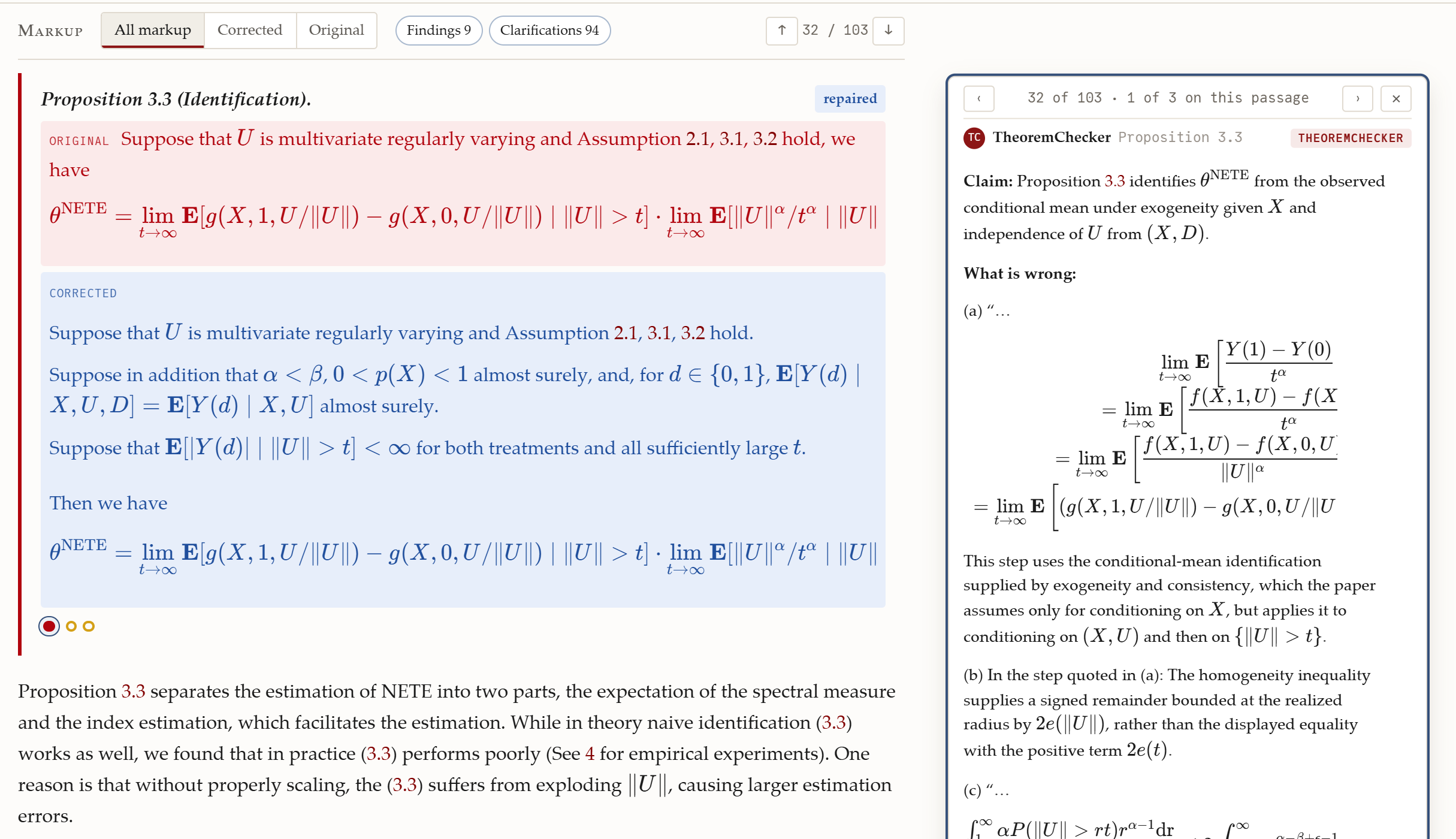}
\caption{Proposition~3.3 of the case-study paper on its page in the \TC{} viewer. The printed statement is shown in red and the confirmed repair in blue; the card on the right gives the reviewer's finding, quoting each step of the proof that fails. The dots under the statement open the other notes on this passage, here two clarifications.}
\label{fig:ui-finding}
\end{figure}

\paragraph{Repairing the identification result.}
Proposition~3.3 claims, under exogeneity given the covariates $X$, $(Y(1),Y(0))\perp D\mid X$, and independence of $U$ from $(X,D)$, that
\[
\theta^{\mathrm{NETE}}
= \lim_{t\to\infty}\mathbb E\bigl[g(X,1,U/\|U\|)-g(X,0,U/\|U\|)\bigm|\|U\|>t\bigr]
\cdot\lim_{t\to\infty}\mathbb E\bigl[\|U\|^{\alpha}/t^{\alpha}\bigm|\|U\|>t\bigr],
\]
where $g$ is the angular part of the outcome regression and $\alpha$ its homogeneity exponent.
The first step of the proof replaces $\mathbb E[Y(d)\mid X,U]$ by the observed regression $f(X,d,U)=\mathbb E[Y\mid X,D=d,U]$. That step needs exogeneity given $(X,U)$, which the two assumptions do not imply.
The reviewer gave a counterexample within the paper's setting.
Let $X\sim\mathrm{Unif}[0,1]$, $D\sim\mathrm{Ber}(1/2)$ and $V\sim\mathrm{Unif}[0,1]$ be independent, let $W$ be independent of them with $\mathbb P(W>w)=w^{-3}$ for $w\ge1$, and set $U=W(V,1-V)$ and $\alpha=1$.
With $T(w)=(1-w^{-3})^{-1/3}$, put $Y(1)=DW+(1-D)T(W)$ and $Y(0)=0$.
Since $T(W)$ and $W$ have the same law, $(Y(1),Y(0))\perp D\mid X$; homogeneity holds exactly with $g(x,1,s)=1$ and $g(x,0,s)=0$.
The right-hand side above equals $3/2$, while $\theta^{\mathrm{NETE}}=3/4$.
The reviewer also found three smaller gaps in the same proof: it chooses $\epsilon>0$ with $\alpha-\beta+\epsilon<0$, which requires $\alpha<\beta$; it states the radial limit as $\alpha/(\beta-\alpha)$ instead of $\beta/(\beta-\alpha)$; and it infers uniform integrability from a uniform bound on first moments.
The repair we accepted keeps the identification formula and adds the conditions the proof uses: $\alpha<\beta$, $0<p(X)<1$, conditional mean exogeneity $\mathbb E[Y(d)\mid X,U,D]=\mathbb E[Y(d)\mid X,U]$, and finite conditional tail means of $|Y(d)|$.
Figure~\ref{fig:ui-finding} shows the finding as the author reads it on the paper's page.

\paragraph{Repairing the rates.}
The repairs in the non-asymptotic analysis address errors of a different kind.
In Lemma~A.1, the selection rule of the adaptive Hill estimator compares candidates up to rank $n$, which requires the order statistic of rank $n+1$; the statistic in the proof differs from the one in Algorithm~1; and the density used in the proof has the outcome exponent $\alpha$ where the tail index $\beta$ belongs. The repair states the rate for a full-sample Hill estimator with prefix selection. Its exponent becomes $1/3$ for $\beta<1$, and it keeps a $\log n$ factor at $\beta=1$.
In Lemma~A.2, the rate $t^{\beta/2}n^{-1/2}$ of the sampling term needs a bounded variance of the normalized score, which the assumptions do not give. The reviewer took symmetric stable noise of index $3/2$: the error of the doubly robust estimator then has scale $n^{-2/15}$, while the printed bound is of order $n^{-1/5}$, so the probability that the bound holds tends to zero.
The repaired lemma replaces the square-root term by a truncated-variance bound, replaces $e(t)$ by its tail supremum $\sup_{r\ge t}e(r)$, and keeps a logarithmic factor in the angular bias at $\beta=1$.
The rate of the angular Wasserstein term is taken from a cited result~\cite{zhang2023wasserstein} and enters as an explicit assumption.

\begin{figure}[t]
\centering
\includegraphics[width=\textwidth]{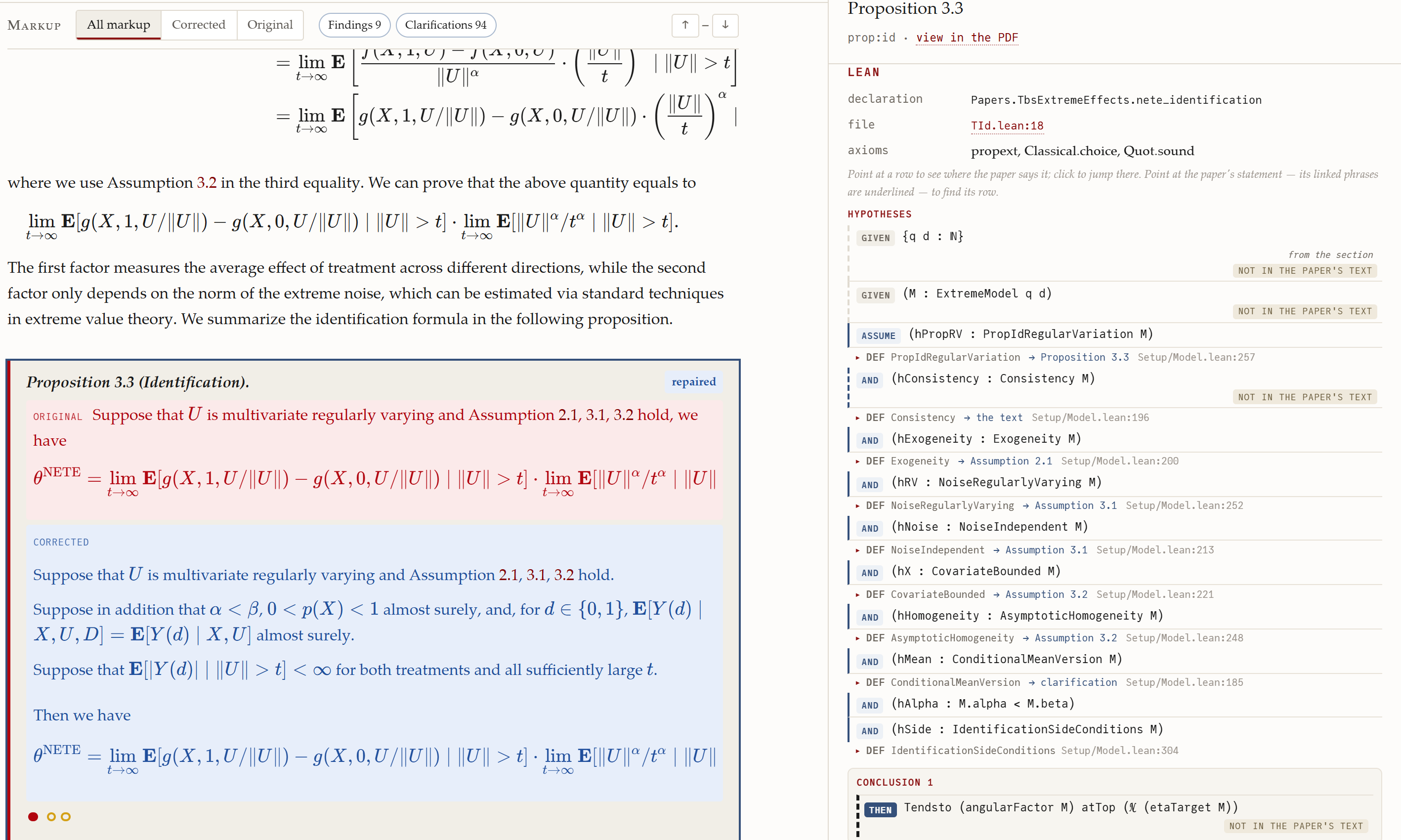}
\caption{The Lean statement of Proposition~3.3 next to the revised statement. Each hypothesis of the Lean theorem links to the assumption of the paper it formalizes (Assumption~2.1, 3.1 or 3.2, a clarification, or the proposition itself) and to its definition in the Lean files; pointing at a row highlights the matching phrase of the paper.}
\label{fig:ui-lean}
\end{figure}

\paragraph{The Lean code.}
The Lean development of the paper has 136 files and about 26{,}000 lines.
It also uses 13 modules (about 2{,}500 lines) from the Foundations tier of the authors' toolkit, among them Chernoff bounds, conditioning on events, and R\'enyi's representation of exponential order statistics, which the proof of Lemma~A.1 uses.
Figure~\ref{fig:ui-lean} shows the link between the paper and the Lean code for Proposition~3.3.
Each hypothesis of the Lean theorem \texttt{nete\_identification} is listed with the assumption of the paper it formalizes, so the author can confirm the formalization assumption by assumption without reading the Lean proofs.
Rows that state something the paper leaves implicit, such as the ambient model, are marked as not in the paper's text.

\paragraph{Shared papers.}
We shared the verified versions of this paper and two other papers of ours on the public website. 
In the paper on optimal transport, the run proved all twelve results in Lean as printed. In the paper on neural causal models, it proved 20 of 29 results and refuted two results for which no repair keeps the claim; neither is a main result of the paper.

\section{Usage and Current Status}

\TC{} is used through a coding agent. The author asks Claude Code or Codex to install it from the repository, chooses the domain libraries, and then asks the agent to formalize a paper.
The agent runs the pipeline, handles the routine steps, and asks the author on the web page whenever a decision is needed.

Both the natural language review and the formalization proceed end to end without supervision, apart from the decision points described in Sections~\ref{sec:pipeline} and~\ref{sec:author}: the interpretation of an ambiguous statement, a repair that would {substantively change} a main result, the plan review, a proposed revision of a paper statement, and the promotions into the toolkit.
An author who wishes to hand these decisions to the agent as well can start the run in a fully automatic mode. Each decision then follows the agent's recommendation and is listed on the web page, where the author approves or rejects it afterwards.
We suggest this mode only for a first pass, since a decision made on the author's behalf may differ from what the author intends.

Formal verification of research papers has been rare because it is costly.
\TC{} aims to make it a routine step of writing a paper rather than a separate project, so that an author can submit or publish a paper together with Lean proofs of its results.
As more authors share verified papers, new work can build on results that are already checked, and the Lean code of one paper becomes the starting point of the next.
We see this as a step toward research in which authors verify the mathematics of their own papers as they write them.


\begin{thebibliography}{99}
\setlength{\itemsep}{0pt}
\footnotesize
\bibitem{lean4}
L.~de~Moura and S.~Ullrich.
\newblock The Lean 4 theorem prover and programming language.
\newblock In Automated Deduction -- CADE 28, pages 625--635, 2021.

\bibitem{mathlib}
The mathlib Community.
\newblock The Lean mathematical library.
\newblock In Proceedings of the 9th ACM SIGPLAN International Conference on Certified Programs and Proofs (CPP), pages 367--381, 2020.

\bibitem{osverify}
F.~Nie, H.~Ye, S.~Liu, Z.~Wang, S.~Ermon, and J.~Zou.
\newblock One-sided verification under limited expert review.
\newblock arXiv:2610.18089, 2026.

\bibitem{rethlas}
H.~Ju, G.~Gao, J.~Jiang, B.~Wu, Z.~Sun, S.~Liu, L.~Chen, Y.~Wang, Y.~Wang, Z.~Wang, W.~He, P.~Wu, L.~Xiao, R.~Liu, B.~Dai, and B.~Dong.
\newblock Automated conjecture resolution with formal verification.
\newblock arXiv:2604.03789, 2026.

\bibitem{leanmarathon}
Y.~Zhang, Y.~Sun, T.~Suzuki, J.~D. Lee, and F.~Liu.
\newblock LeanMarathon: Toward reliable AI co-mathematicians through long-horizon Lean autoformalization.
\newblock arXiv:2606.05400, 2026.

\bibitem{leanflow}
L.~Milikic, S.~Guilloud, K.~Nguyen, and V.~Kuncak.
\newblock LeanFlow: A case study in workflow-driven Lean autoformalization.
\newblock arXiv:2607.20503, 2026.

\bibitem{causalsmith}
J.~Tan and V.~Syrgkanis.
\newblock CausalSmith: A formally grounded, self-improving agentic framework for automated research in causal inference.
\newblock arXiv:2607.22511, 2026.

\bibitem{statlean}
The StatLean contributors.
\newblock Stat-Lean: a Lean library for statistics.
\newblock \url{https://github.com/StatLean/Stat-Lean}.

\bibitem{foml}
S.~Sonoda, K.~Kasaura, Y.~Mizuno, K.~Tsukamoto, and N.~Onda.
\newblock Lean formalization of generalization error bound by Rademacher complexity and Dudley's entropy integral.
\newblock arXiv:2503.19605, 2025.

\bibitem{optlib}
C.~Li, S.~Xu, C.~Sun, L.~Zhou, and Z.~Wen.
\newblock Formalization of optimality conditions for smooth constrained optimization problems.
\newblock arXiv:2503.18821, 2025.

\bibitem{loogle}
J.~Breitner.
\newblock Loogle: a search engine for Lean and Mathlib.
\newblock \url{https://loogle.lean-lang.org}.

\bibitem{leansearch}
G.~Gao, H.~Ju, J.~Jiang, Z.~Qin, and B.~Dong.
\newblock A semantic search engine for Mathlib4.
\newblock arXiv:2403.13310, 2024.

\bibitem{statlean}
The StatLean contributors.
\newblock Stat-Lean: a Lean library for statistics.
\newblock \url{https://github.com/StatLean/Stat-Lean}.

\bibitem{foml}
S.~Sonoda, K.~Kasaura, Y.~Mizuno, K.~Tsukamoto, and N.~Onda.
\newblock Lean formalization of generalization error bound by Rademacher complexity and Dudley's entropy integral.
\newblock arXiv:2503.19605, 2025.

\bibitem{optlib}
C.~Li, S.~Xu, C.~Sun, L.~Zhou, and Z.~Wen.
\newblock Formalization of optimality conditions for smooth constrained optimization problems.
\newblock arXiv:2503.18821, 2025.

\bibitem{loogle}
J.~Breitner.
\newblock Loogle: a search engine for Lean and Mathlib.
\newblock \url{https://loogle.lean-lang.org}.

\bibitem{leansearch}
G.~Gao, H.~Ju, J.~Jiang, Z.~Qin, and B.~Dong.
\newblock A semantic search engine for Mathlib4.
\newblock arXiv:2403.13310, 2024.

\bibitem{zhang2023wasserstein}
X.~Zhang, J.~Blanchet, Y.~Marzouk, V.~A. Nguyen, and S.~Wang.
\newblock Wasserstein-based minimax estimation of dependence in multivariate regularly varying extremes.
\newblock arXiv:2312.09862, 2023.
\end{thebibliography}
\end{document}